\documentclass[a4paper, 10pt, conference]{ieeeconf}      

\IEEEoverridecommandlockouts                              

\usepackage{graphics} 
\usepackage{epsfig} 
\usepackage{mathptmx} 
\usepackage{times} 
\usepackage{amsmath} 
\usepackage{amssymb}  

\usepackage{subfigure} 
\usepackage{booktabs}
\usepackage{bm}
\usepackage{multirow}
\usepackage{ctable}
\usepackage{threeparttable}
\usepackage{booktabs}
\usepackage{multirow}

\usepackage{flushend}

\title{\LARGE \bf
	Coarse-To-Fine Visual Localization Using Semantic Compact Map
}

\author{Ziwei Liao$^{1,3}$, Jieqi Shi$^{2}$, Xianyu Qi$^{1}$, Xiaoyu Zhang$^{1}$, Wei Wang$^{1,*}$, Yijia He$^3$, Ran Wei$^4$, and Xiao Liu$^3$ 
\thanks{* Corresponding author.} %
\thanks{$^{1}$ Institute of Robotics, Beihang University (BUAA) {\tt\small \{liaoziwei, wangweilab\}@buaa.edu.cn}} %
\thanks{$^{2}$ Hong Kong University of Science and Technology}
\thanks{$^{3}$ Megvii Inc. (Face++), China}
\thanks{$^{4}$ Beijing Evolver Robotics Technology Co., Ltd., China}
}

\begin{document}

\maketitle
\thispagestyle{empty}
\pagestyle{empty}

\begin{abstract}
        Robust visual localization for urban vehicles remains challenging and unsolved. The limitation of computation efficiency and memory size has made it harder for large-scale applications. 
        Since semantic information serves as a stable and compact representation of the environment, we propose a coarse-to-fine localization system based on a semantic compact map. Pole-like objects are stored in the compact map, then are extracted from semantically segmented images as observations. Localization is performed by a particle filter, followed by a pose alignment module decoupling translation and rotation to achieve better accuracy. We evaluate our system both on synthetic and realistic datasets and compare it with two baselines, a state-of-art semantic feature-based system, and a traditional SIFT feature-based system. Experiments demonstrate that even with a significantly small map, such as a 10 KB map for a 3.7 km long trajectory, our system provides a comparable accuracy with the baselines. 
        \end{abstract}
        
        \section{Introduction}
        
        Robust visual localization for autonomous vehicles has gained wide attention recently. The traditional visual localization pipeline builds up a map from texture features \cite{svarm2016city,sattler2016efficient} of the environment or uses image retrieval techniques \cite{torii201524,arandjelovic2016netvlad} for camera state estimation. To ensure robustness in urban scenarios, researchers must carefully design features \cite{lowe2004distinctive,bay2006surf} unaffected by viewpoints and highly dynamic objects. However, almost all popular features still cannot adapt well to drastic changes in appearance \cite{aanaes2012interesting} or rely too much on data used for training \cite{yi2016lift}. Besides, the complex descriptor makes the map storage too large and limits computation efficiency.
        
        With the development of deep-learning techniques, many researchers have turned to semantic segmentation for an advanced representation of the environment. Toft et al. \cite{toft2018semantic} use semantic information to assign a weight to points in the RANSAC process, which improves the robustness during the PnP process. Schonberger et al. \cite{Schonberger2018} learn a descriptor that encodes both 3D geometry and semantic information. 
        These methods treat semantic information as auxiliary information for traditional features. They may improve robustness and accuracy, but can not fulfill our need for using as little storage as possible.
        
        \begin{figure}[t]
                \begin{center}
                        \includegraphics[width=0.9\linewidth]{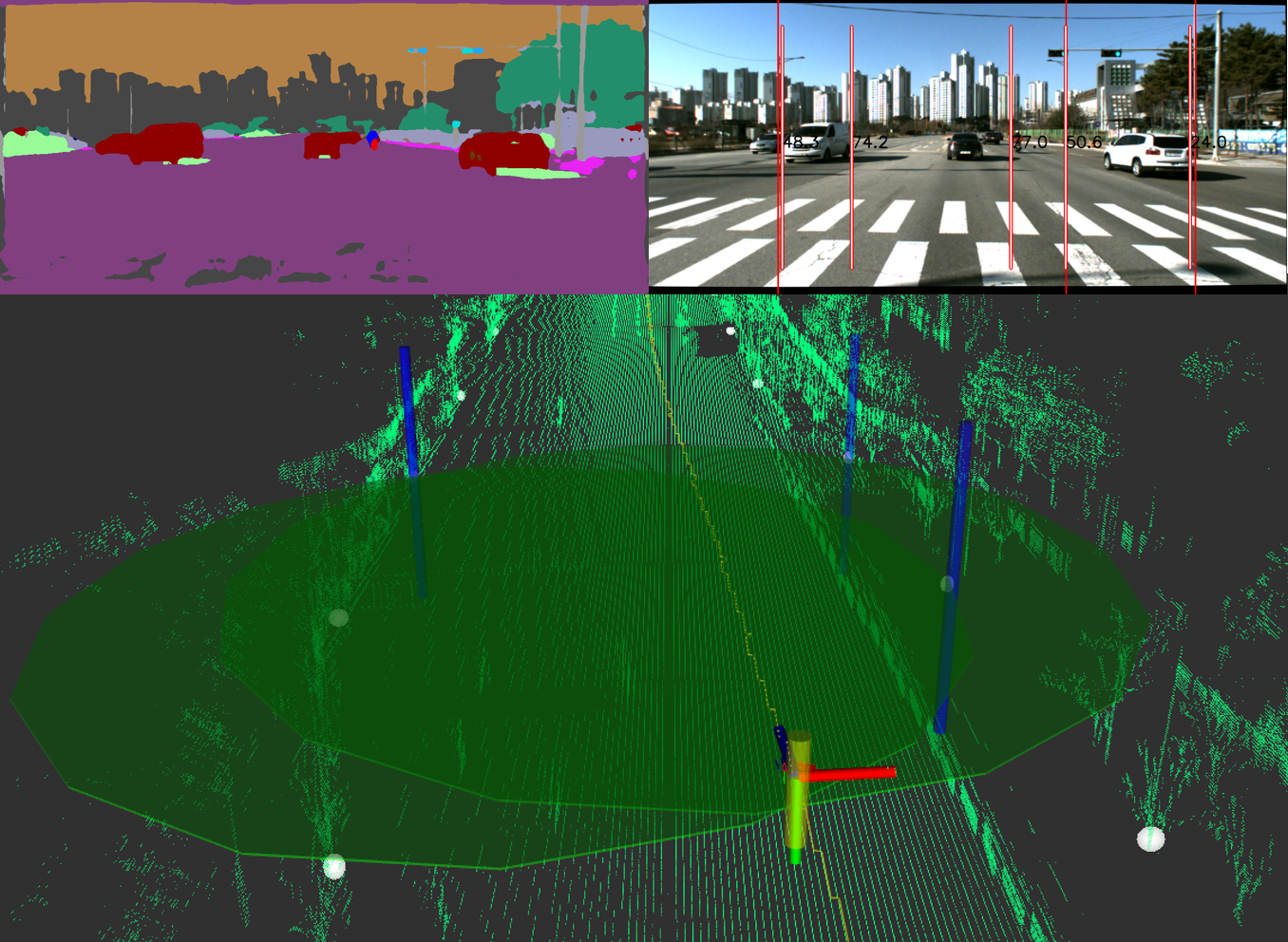}
                        
                \end{center}
                \caption{ This figure demonstrates the fine positioning process. In the current frame, the image on the upper left is the result of semantic segmentation. In the image on the upper right, the extracted poles are marked with a long red line. The poles projected from the compact map are marked with a short white line, and their distances are marked in meters with black numbers.
                        In the figure below, the poles with successful data associations are marked with a blue cylinder. Two green circles are calculated based on the positions of poles and observation angles. Then the camera translation is estimated as the intersection of the circles. This method solves the camera state decoupling translation and rotation. We recommend viewing in color.}
                \label{fig:intro}
        \end{figure}
        
        Some researchers have proposed systems that require only semantic information. Toft et al. \cite{toft2017long} store a semantic point cloud and extract semantic boundaries on semantically segmented images to build a cost map for the optimization framework. 
        Most of the methods we mentioned above follow a typical SLAM method, with a graph optimization process optimizing the results. However, graph optimization is not efficient enough in large-scale environments, requiring a complex marginalization algorithm to manage keyframes and simplify the graph. Stenborg et al. \cite{stenborg2018long} form a particle filter system, using the match consistency between the semantic point cloud and semantically segmented images as particle weights. 
        The storage for every semantic landmark has dramatically reduced compared with the complex feature descriptor. However, the storage needed for the point cloud remains large as a massive number of semantic points are required, making the system hard to run online. 
        
        Some researchers have proposed compact maps that store objects instead of semantic points, which requires much less storage space. Weng et al. \cite{weng2018pole} consider poles extracted from lidar observations for scan-matching and fuse an RTK GPS to achieve high accuracy. Spangenberg et al. \cite{spangenberg2016pole} consider tree trunks and use a particle filter with a stereo camera. They all contribute a lot to the development of semantic compact maps but ask for some expensive equipment for object detection. 
        
        
        
        
        
         
        To address the problem of making map storage as small as possible, we choose object-level landmarks to form a semantic compact map. We believe pole-like objects, such as traffic light, lamp, and tree trunks, are everyday objects in urban sceneries and can be used as reliable landmarks for localization. Differed from \cite{spangenberg2016pole}, which extracts tree trunks from stereo images, we derive new observation models for pole-like objects and propose a novel method to extract object observations from semantically segmented images directly. 
            
        To consider both efficiency and accuracy, we propose a coarse-to-fine localization system. The coarse localization system forms a particle filter for efficient state estimation. Differed from \cite{spangenberg2016pole}, we propose the pose alignment module to adjust the pose from the particle filter to a finer position. With accurate landmarks in the compact map, we can use a geometry solution that decouples translation and rotation \cite{Sugihara1988,Kanatani2001} to achieve higher accuracy. 
            
        We compare the system performance with two baselines using synthetic and realistic datasets covering urban, suburban, bridge, and highway scenarios. We aim to show that even with a significantly small map, the system could achieve comparable accuracy with the state-of-art semantic localization system and traditional point-cloud based systems.
        
        In summary, we make the following contributions in this work:
        \begin{itemize}
        
                \item[(1)] We propose a coarse-to-fine localization system using a particle filter for autonomous driving scenarios based on a semantic compact map.
                \item[(2)] We propose an observation model for pole-like objects, and a method to extract them from semantically segmented monocular images.
                \item[(3)] We introduce a novel pose alignment module to decouple translation from the rotation to estimate an accurate pose.
                \item[(4)] We demonstrate our methods on public datasets and achieve comparable accuracy compared with two baselines with a much smaller map.	
        \end{itemize}
        
        
        \section{System Framework}
        
        \begin{figure}[t]
                \begin{center}
                        \includegraphics[width=0.9\linewidth]{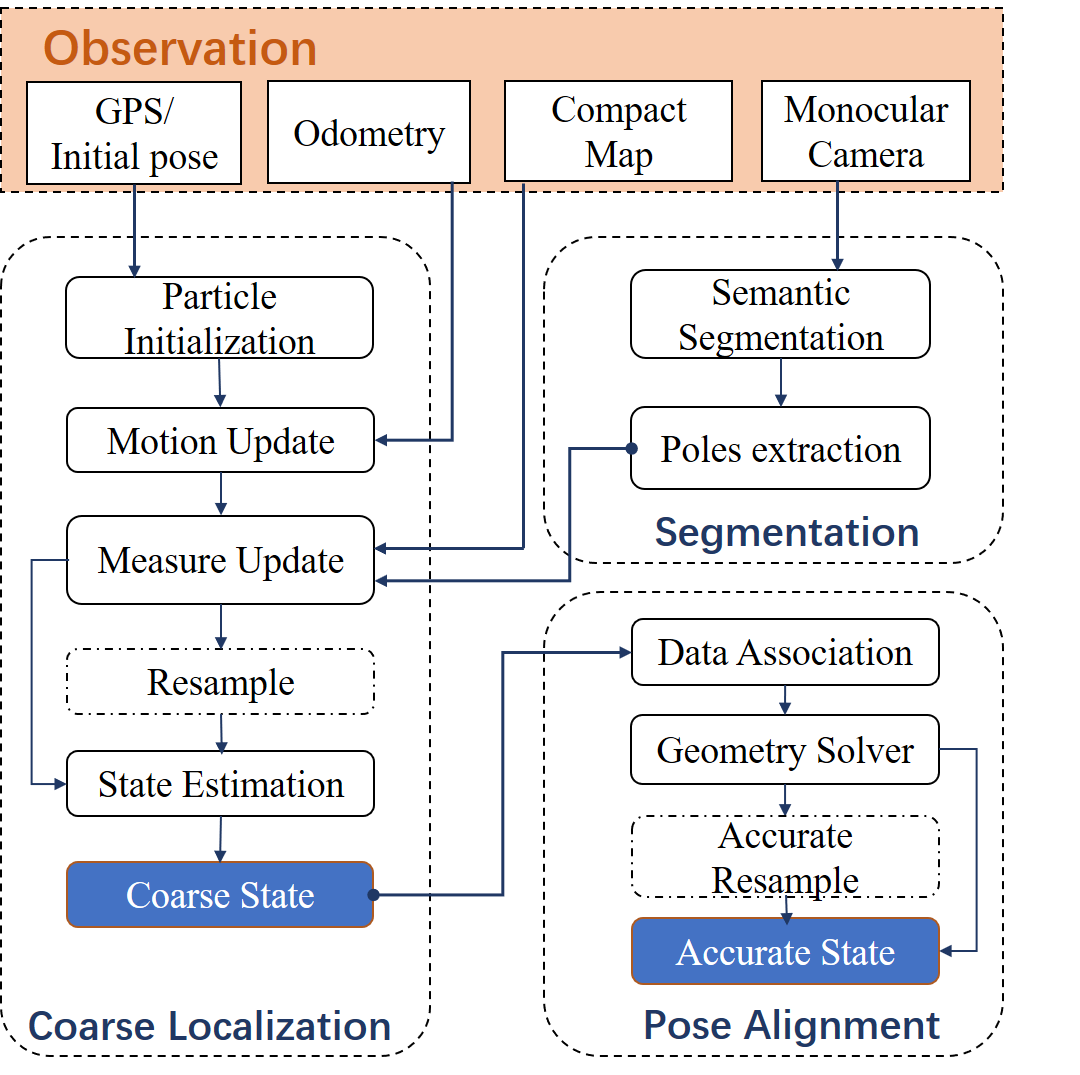}
                        
                \end{center}
                \caption{System Framework. The proposed system takes initial pose, odometry, compact map and monocular camera as input. It consists of three main components. The Segmentation module extracts poles observations from monocular camera images. The Coarse Localization module updates camera state using a particle filter. The Pose Alignment module uses a novel method to get an accurate state. The Resample process only activates when necessary.
                }
                \label{fig:sys}
                
        \end{figure}
        \subsection{Overview}
        Our coarse-to-fine localization system consists of three dependent modules: Segmentation, Coarse Localization, and Pose Alignment, as in Fig. \ref{fig:sys}. 
        
        The whole system is initialized using GPS or other rough localization systems like \cite{Ardeshir2014} only once. After localizing the camera's initial pose, we will operate the segmentation module and the coarse localization module in parallel. The segmentation module outputs poles extracted from the semantic segmentation algorithm using CNN models. In the meantime, the camera pose is calculated roughly in the Motion Update step of the coarse localization module via the odometry provided. The compact map is then projected to the camera to establish a data association between poles extracted by semantic segmentation and poles saved in the compact map. After every particle is assigned a weight through the Measurement Update, resampling is done when it is necessary. Then, poses are estimated by state estimation.
        
        The pose alignment module is our main contribution, which is designed to optimize the coarse localization module's pose and can be closed to save time or operated at a given frequency, such as every five frames. This module uses a geometry-based pose solver and decouples translation and rotation \cite{Sugihara1988} to solve an accurate translation of the camera. Then rotation is optimized with translation fixed. The results of this module will be evaluated. If it is decided to be better than coarse localization, another resample step will follow before outputting the final result.
        
        \subsection{Pole Representation}
        \label{sec_pole_representation}
        We assume that the road surface is flat, and the camera is parallel to the road. Though it is a strong assumption, we will prove in the experiments that it is reasonable in a wide range of realistic autonomous driving conditions, e.g., urban, suburban, bridge, and highways. 
        
        Based on the assumption above, we model all semantic objects in our map as infinite poles vertical to the ground. Such representation makes our map further compressed and light-weight. Since we do not consider their height, we can record landmarks in a two-dimensional state vector. Therefore, the compact map is composed of $M$ poles with positions and semantic labels as $L =\left \{ \left\langle p_i, a_i \right\rangle \right \}_{i=1}^M$, where $p_i=[x_i,y_i]^T$ and $a_i \in$ \{Pole, Lamp, Tree Trunk, ...\}.  
        
        Given a particle pose $m$ with pose $x_m$ and poles in the map, projections of poles on the image plane of $m$ are infinite lines vertical to the x-coordinate, shown as long red lines in Fig.1. 
        We record the x coordinate $u$ and semantic label $a$ of each poles as projections $\hat{l}_m = \left\{\left\langle \hat{u}_m^k, \hat{a}_m^k\right\rangle \right\}_{k=1}^M$. Later, we will compare them with the poles observations from semantically segmented images to solve data associations and calculate the camera pose.

        \begin{figure}[t]
                \begin{center}
                        
                        \subfigure[semantic segmentation]{
                                \begin{minipage}[t]{0.5\linewidth}
                                        \centering
                                        \includegraphics[width=1.0\linewidth]{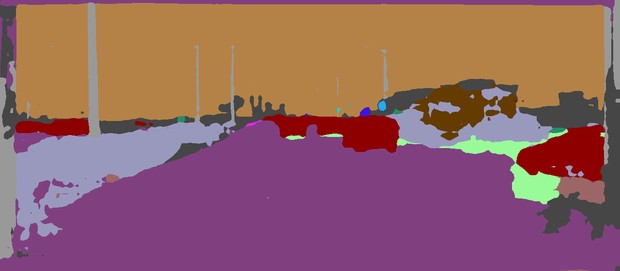}  
                                \end{minipage}%
                        }%
                        \subfigure[binarized image]{
                                \begin{minipage}[t]{0.5\linewidth}
                                        \centering
                                        \includegraphics[width=1.0\linewidth]{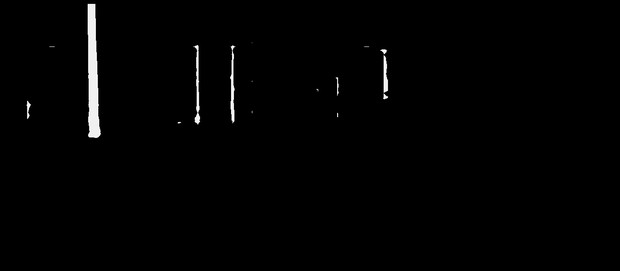} 
                                \end{minipage}%
                        }%
                        \quad
                        \subfigure[middle result]{
                                \begin{minipage}[t]{0.5\linewidth}
                                        \centering
                                        \includegraphics[width=1.0\linewidth]{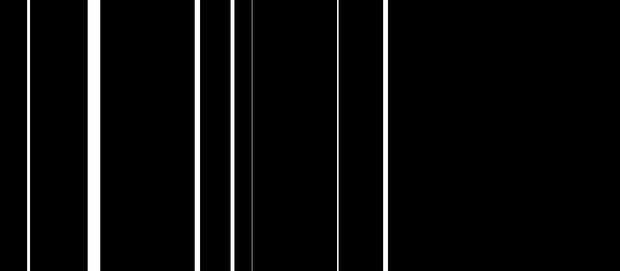} 
                                \end{minipage}%
                        }%
                        \subfigure[pole extraction]{
                                \begin{minipage}[t]{0.5\linewidth}
                                        \centering
                                        \includegraphics[width=1.0\linewidth]{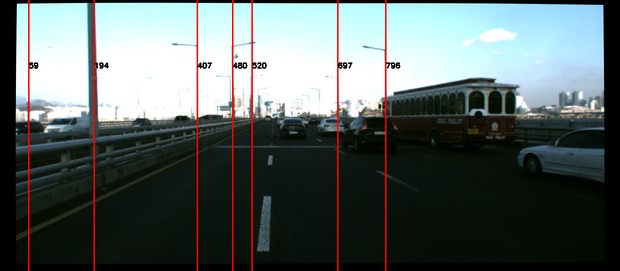}
                                \end{minipage}%
                        }%

                \end{center}
                \caption{Pole Extraction and Modeling. We extract pole-like objects from semantically segmented images and model their observations as lines vertical to the x-coordinate. Black numbers in (d) show the x-coordinate values. Please see Section \ref{sec_pole_representation} for the detail.}
                \label{fig:pole}
                
        \end{figure}
        
        \subsection{Pole Extraction}
        In the segmentation module, we input RGB images and process them using semantic segmentation methods \cite{yu2018bisenet}. For each frame $F_t$, we filter output of the network to extract $n_t$ poles, and store their x coordinates and semantic labels as: $\bar{l}_t = \left\{\left \langle \bar{u}_t^i, \bar{a}_t^i \right\rangle \right\}_{i=1}^{n_t}$. We divide the filtering algorithm into three steps, as shown in Fig. \ref{fig:pole}.
        
        \begin{itemize}
                \item[(1)] Binarize the segmentation image, marking poles and other regions as 1 and 0.
                \item[(2)] Count how many pixels are labeled as poles in each column and filter out columns that do not reach the threshold $c_1$ (We use $c_1=60$ in experiments). Successive columns that have not been filtered out will be grouped.
                \item[(3)] Calculate the width of each column group. If the width is in range $[c_2,c_3]$ (We use $c_2=1,c_3=15$ in experiments), we will extract a pole located in the middle of this group.
        \end{itemize}

        \section{Coarse Localization}
        
        The Coarse Localization module is based on a particle filter similar to \cite{spangenberg2016pole}. We design the filter based on our proposed pole-like observations from monocular images, instead of tree-truncks from stereo images in \cite{spangenberg2016pole}. We frequently sample several particles at each pose and update the camera's pose by fusing these particles' poses. For every frame the module receives, it will estimate the camera pose after motion update, measurement update, resampling, and state estimation. 
        
        \subsection{Motion Update}
        
        We use UTM(Universal Transverse Mercator) coordinate, which takes East and North as two vertical axes and records position together with orientation, to describe camera pose. We note the state of i-th particle in the t-th frame as $p_t^i=\big[E_t^i\ N_t^i\ \psi_t^i\big]^T$.
        
        During the system's initialization, we read in GPS data and its uncertainty to produce an initial distribution of particles. When a new frame comes, we will update the state of every particle via its linear velocity ${v_t}$ and angular velocity ${\omega_t}$ from the odometry:
        \begin{equation}
        \begin{aligned}
        {\scriptscriptstyle
                \left( \begin{array}{l}{E_{t+1}} \\ {N_{t+1}} \\ {\psi_{t+1}}\end{array}\right)=\left( \begin{array}{l}{E_{t}} \\ {N_{t}} \\ {\psi_{t}} \end{array}\right)+\left( \begin{array}{c}{-\frac{\hat{v_t}}{\hat{\omega_t}} \sin \psi_t+\frac{\hat{v_t}}{\hat{\omega_t}} \sin (\psi_t+\hat{\omega_t} \Delta t)} \\ {\frac{\hat{v_t}}{\hat{\omega_t}} \cos \psi_t-\frac{\hat{v_t}}{\hat{\omega_t}} \cos (\psi_t+\hat{\omega_t} \Delta t)} \\ {\hat{\omega_t} \Delta t+\hat{\gamma_t} \Delta t}\end{array}\right)
        }
        \end{aligned},
        \end{equation}
        where $\hat{v_t}=v_t+\varepsilon_{\alpha_1 |v_t|+\alpha_2 |w_t|}, \hat{\omega_t}=\omega_t+\varepsilon_{\alpha_3 |v_t|+\alpha_4 |\omega_t|}$, and $\varepsilon_\sigma$ obeys the Gauss distribution with a mean of 0 and a standard deviation of $\sigma$.
        To avoid degeneracy, we assume the vehicle performs a rotation $\hat{\gamma_t}=\varepsilon_{\alpha_5 |v_t| + \alpha_6 |\omega_t|}$ after it arrives the final pose (see \cite{thrun2005probabilistic}, p.129). The parameters $\alpha_{1},...,\alpha_{6}$ are decided by the configuration of vehicles or robots we use.
        
        \subsection{Measurement Update} \label{sec_measurement_update}
        
        With poles extracted in the Segmentation Module and the compact map, we update the state of each particle independently.
        
        \subsubsection{Data Association}
        
        Assuming that in the current frame $F_t$, there are $n_t$ poles observations and $M$ poles in our map. As mentioned in Section \ref{sec_pole_representation}, we note the poles observations in the $t$-th frame as $\bar{l}_t = \left\{\left \langle \bar{u}_t^i, \bar{a}_t^i \right\rangle \right\}_{i=1}^{n_t}$ and the projection of poles in the map to the state of particle $m$ as $\hat{l}_m = \left\{\left\langle \hat{u}_m^k, \hat{a}_m^k\right\rangle \right\}_{k=1}^M$. 
        
        Considering a particle $m$ in t-th frame, we omit $t,m$ for clarity and write poles observations as $\bar{l} = \left\{\left \langle \bar{u}^i, \bar{a}^i \right\rangle \right\}_{i=1}^{n}$ and poles projections as $\hat{l} = \left\{\left\langle \hat{u}^k, \hat{a}^k\right\rangle \right\}_{k=1}^M$.
        To estimate the particle state, we need to establish the data associations between $\bar{l}$ and $\hat{l}$. The data association problem can be modeled as an Optimal Mapping problem \cite{jonker1987shortest}. Taken that we have already found a mapping $\theta :\left\{1, \ldots, n \right\} \rightarrow\left\{0,1, \ldots, M \right\}$, where $\theta(i) = 0$ indicates that the i-th pole observation does not belong to any pole in our map. We define a loss function $\mathrm{e}(i, k)$ as the distance between the observation $i$ and the projection of pole $k$ on the image plane:
        \begin{equation}
                \mathrm{e}(i, k) =  |  \bar{u}^i - \hat{u}^k | \label{eq-diff}.
                \end{equation}
        Assuming the pose of pole $k$ in the map is $p_k=[x_k,y_k]^T$, its pose in camera coordinate is $p_k^{'}=[x_k^{'},y_k^{'}]^T$, and the pose of particle m is $x_m=\big[E_m\ N_m\ \psi_m\big]^T$, the projection model is:
        \begin{equation}
                \begin{gathered}
                        \begin{bmatrix} x_k^{'} \\ y_k^{'} \\ 1  \end{bmatrix} 
                        = 
                        \begin{bmatrix} 
                                 \cos \psi_m  & -\sin \psi_m & E_m \\
                                 \sin \psi_m & \cos \psi_m  & N_m \\
                                 0 & 0 & 1
                        \end{bmatrix}^{-1}
                        \begin{bmatrix} x_k \\ y_k \\ 1  \end{bmatrix} 
                \end{gathered}.
        \end{equation}
        \begin{equation}
                \hat{u}^k = \frac {x_k^{'}f_x} {y_k^{'}} +c_x,
        \end{equation}
        $f_x, c_x$ is the camera intrinsic. The total loss $E$ is defined by adding together all distances between the observation $\bar{u}$ and its corresponding projection $\hat{u}$:
        $ E = \sum_{1 \leq i \leq n, k = \theta(i)} \mathrm{e}(i, k)$.
        By minimizing the total loss of $E$, we can acquire an optimum estimation of $\hat{\theta}$ of the real mapping $\theta$ between observations and poles in the map. 
        
        \subsubsection{Weight Calculation}
        Every particle $m$ represents a possible camera state. We first assign a weight to it based on poles' observations and projections. We consider every pole observation $\bar{l}_i \in \bar{l}$ in two categories:
        \begin{itemize}
                \item[(1)] The observation $\bar{l}_i$ is mapped to an established pole in the map $\left(k=\hat{\theta}(i) \textgreater 0\right)$. 
                Similar to \cite{spangenberg2016pole}, we define that
                \begin{equation}
                \mathrm{d}(i, k) = \beta_{s} \cdot \mathrm{e}(i, k)^2 \cdot S^{-1},
                \end{equation}
                where $\beta_s$ is of a positive correlation with the distance between pole and camera in 3D space. We assume that closer poles have higher weight contributing to the cost. Differed from \cite{spangenberg2016pole}, we model the pole distance instead of the pole width as weight, as we find that the pole width is hard to estimate because of segmentation noise. $S^{-1}$ is used to model the uncertainty of observations and is decided by the pole extraction module. 
                $\mathrm{e}$ is defined in Eq. \ref{eq-diff}. Then, the weight provided by $\bar{l}_i$ can be formulated as
                \begin{equation}
                g\left(\bar{l}_i | \hat{l}_k \right) = \frac{p_{D}}{\kappa\left(\bar{l}_i\right)} \exp \left(-\frac{1}{2}\left(d\left(i, k\right)\right)\right),
                \end{equation}
                where $p_D$ is the possibility that one specific kind of pole is identified and is also related to the pole segmentation model. $\kappa\left(\bar{l}_k\right)$ is intensity of a Poisson clutter process.
                
                \item[(2)] The observation $\bar{l}_i$ is not mapped to an established pole $L_k$ $\left(k=\hat{\theta}(i) = 0\right)$. 
                Under this circumstance, we define
                \begin{equation}
                        g\left(\bar{l}_i | \hat{l}_k \right)=1-p_{D}.
                \end{equation}
                
        \end{itemize}
        
        The weight of $m$-th particle is calculated by considering all the observations: 
        \begin{equation}
        w_m=\prod_{i=1}^{n} g\left(\bar{l}_i | \hat{l}_{\hat{\theta}\left(i\right)}\right). \label{eq-weight}
        \end{equation}
        
        \subsection{State Estimation}
        
        Inspired by the strategy of GMAPPING\cite{grisetti2007improved}, we use the number of efficient particles $N_{eff}$ to judge whether we should do resample. We define that	
        \begin{equation}
        \hat{N_{eff}} \approx  \frac{1}{ \sum_{m=1}^N { { w_m }^2 } }.
        \end{equation}
        Assuming $N$ is the number of particles, we do resample if $\frac{\hat{N_{eff}}}{N}<p_0$ (We use $p_0=0.6$ in experiments). And the camera state is estimated by the weighted average of all particles.

        \section{Pose Alignment}
        Given the result of coarse localization, pose alignment will make use of geometric constraints to decouple translation and rotation. Sugihara et al. \cite{Sugihara1988} and Kanatani et al. \cite{Kanatani2001} describe the theory in detail. To our best knowledge, we first introduce it to solve the visual localization problem for vehicles using pole-like objects. If the pose alignment procedure gives out another camera state that better satisfies our requirements, we take it and run the accurate resample to adjust the particle distribution. 
        
        \subsection{Translation Calculation}
        
        Data association has been solved in Section \ref{sec_measurement_update}. Suppose we have two landmarks $L_1$, $L_2$, their projection to the image plane $l_1 = \left\langle {u}_1, {a}_1\right\rangle $, $l_2 = \left\langle {u}_2, {a}_2\right\rangle $ and the intrinsic matrix of our camera. We can treat the monocular camera as a protractor and calculate horizontal angle $\theta_{12}$:
        \begin{equation}
        \theta_{12} = \arctan(\frac{{u}_1-c_x}{f_x})-\arctan(\frac{{u}_2-c_x}{f_x}), 
        \end{equation}
        where we assume ${u}_1 > {u}_2$. $f_x$, $c_x$ are intrinsic parameters of the camera.
        
        \begin{figure}[t]
                \begin{center}
                        \includegraphics[width=0.65\linewidth]{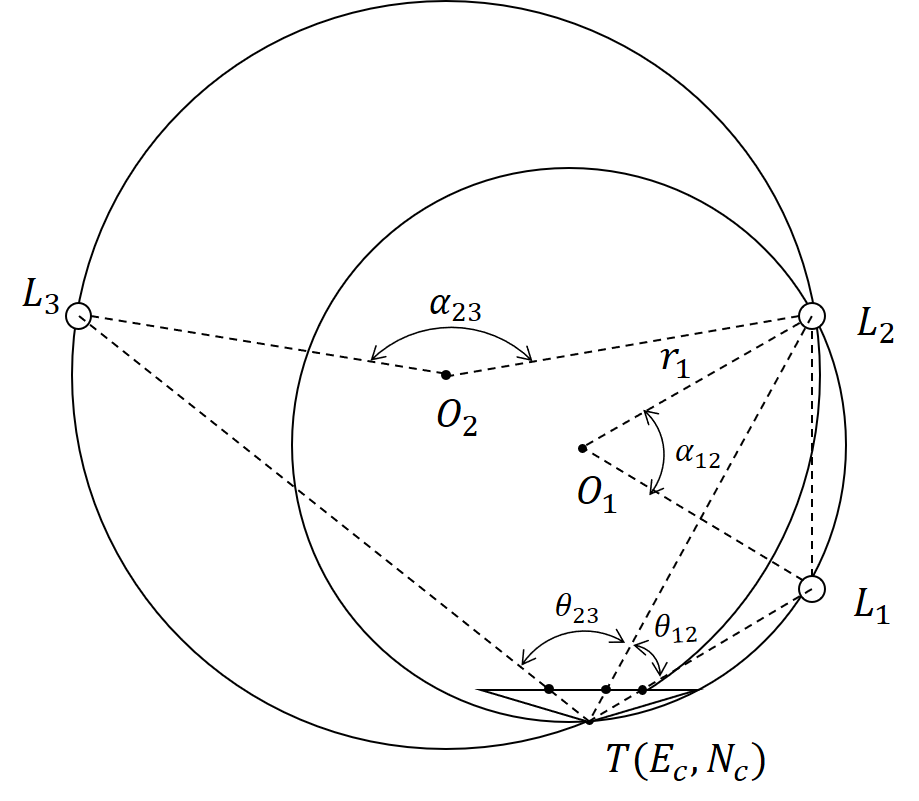}
                \end{center}
                \caption{Translation calculation. Given the landmarks $L_1$, $L_2$, $L_3$, and the horizontal angles $\theta_{12}$, $\theta_{23}$, camera $T$ sits at the intersection of two circles $O_1$ and $O_2$. }
                \label{fig:trans}
        \end{figure}
        
        As shown in Fig. \ref{fig:trans}, $\theta_{12}$ can be seen as an angle in a circular segment and $L_1$, $L_2$ are two points in this circle $O_1$. Based on the circumferential angle theorem, the central angle is two times the corresponding angle in the circular segment. Therefore, we can easily get the radius of this circle:
        \begin{equation}
        r_1=\frac{| L_1L_2 |}{2\sin(\theta_{12})}.
        \end{equation}
        By the same reason, we acquire the circle $O_2$ and its radius $r_2$ through landmark $L_2L_3$ and the viewing angle $\theta_{23}$. $O_1$ and $O_2$ will insert at two points: $L_2$ and the camera $T = (E_c, N_c)$. Since $T$ and $L_2$ are symmetric about $O_1O_2$, we can obtain the camera translation now.
        
        The camera rotation $\psi$ is not used in the calculation of camera translation above, which means that rotation has nothing to do with translation in our method. Thus, we have decoupled translation and rotation. We will demonstrate in the experiments that it helps achieve better accuracy.
        
        \subsection{Rotation Optimization}
        To get a complete pose candidate $\hat{x} = [\hat{T}, \hat{\psi}]^T$, we construct an optimization function to solve the camera rotation $\hat{\psi}$ with its translation $\hat{T}$ fixed:
        \begin{equation}
        \hat{\psi} = \mathop{\arg\min}_{\psi}  \frac{1}{2}\sum_{i=1}^{n} \| \mathrm{e} (i, \hat{\theta} (i))\|_2^2,
        \end{equation}
        where $\mathrm{e}$ is defined in Eq. \ref{eq-diff}. $\hat{\theta}$ is the data association solved in Section \ref{sec_measurement_update}. There is only one parameter to be optimized in this nonlinear optimization problem. We solve this problem using the Gaussian-Newton method. 
        
        \subsection{Evaluation}
        
        During the pose alignment process discussed above, we need three landmarks once to adjust the camera's pose. Suppose that we have $n$ ($n \ge 3$) landmarks successfully associated with observations in the image plane, we obtain $C_n^3$ possible landmark groups and the same number of accurate poses candidates. 
        In realistic applications, there are some possibilities for accurate pose alignment to fail. For example, wrong semantic segmentation results in false observations and data associations. Also, degenerate configurations may occur when two circles, as in Fig. \ref{fig:trans}, are nearly identical. 
        
        We design an evaluation process to keep a robust pose estimation. We calculate a weight for every poses candidates as defined in Eq. \ref{eq-weight}, and sort them by weight. Then, we choose the best pose candidate $x^{*}$ with the highest weight $w_p$. The evaluation process contains two criteria:
        \begin{itemize}
            \item[(1)] $w_p > w_c$, where $w_c$ is the weight of the coarse pose.
            \item[(2)] $d_p < d_0$, where $d_p$ is the distance between the accurate pose $x^{*}$ and the coarse pose (we use $d_0=1m$ in experiments).
        \end{itemize}
        
        
        If both the criteria are met, we treat the pose alignment as successful and activate the accurate resample step. Otherwise, we output the result of coarse localization. When there are less than three successfully observed landmarks in the current frame, we also output coarse localization.
        
        \subsection{Accurate Resample}
        
        If the evaluation process outputs a valid accurate pose, we adjust all particles' probability distribution according to it. We will generate a new Gauss distribution with a mean of $x^{*}$ and a standard deviation of $\sigma_t$ for translation and $\sigma_\theta$ for rotation. Then we sample new particles from this distribution. We assume that the uncertainty of pose alignment is of negative correlation to the pose weight. We define $\sigma_t=(1 - w^{*} ) \beta_t, \sigma_\theta=(1 - w^{*} ) \beta_\theta$, where $\beta_t,\beta_\theta$ reflects the reliability of the pose alignment module (we use $\beta_t=1000$ m, $\beta_\theta=200$ rad in experiments).
        
        Unlike traditional resampling methods that take advantage of existing particles in the Motion Update, the accurate resample step generates distributions from observations and constructed maps. Such approaches are more like a global relocalization technique and reduce the effect of drifts in the long run.
        
        \section{Experiments}
        
        To thoroughly evaluate the performance of our system, we employ both virtual and real datasets. For comparison, we realize a state-of-art semantic localization system that uses a semantic point cloud based on a Particle Filter and a traditional SIFT-UKF system using SIFT features based on an Unscented Kalman Filter (UKF). Both baselines are described in \cite{stenborg2018long}. Our experiments are carried out using a PC with an Intel Core i5-4590 CPU with 3.3GHz, and an Nvidia GeForce GTX TITAN X GPU.
        
        \begin{figure}[t]
                \begin{center}
                        \includegraphics[width=1.0\linewidth]{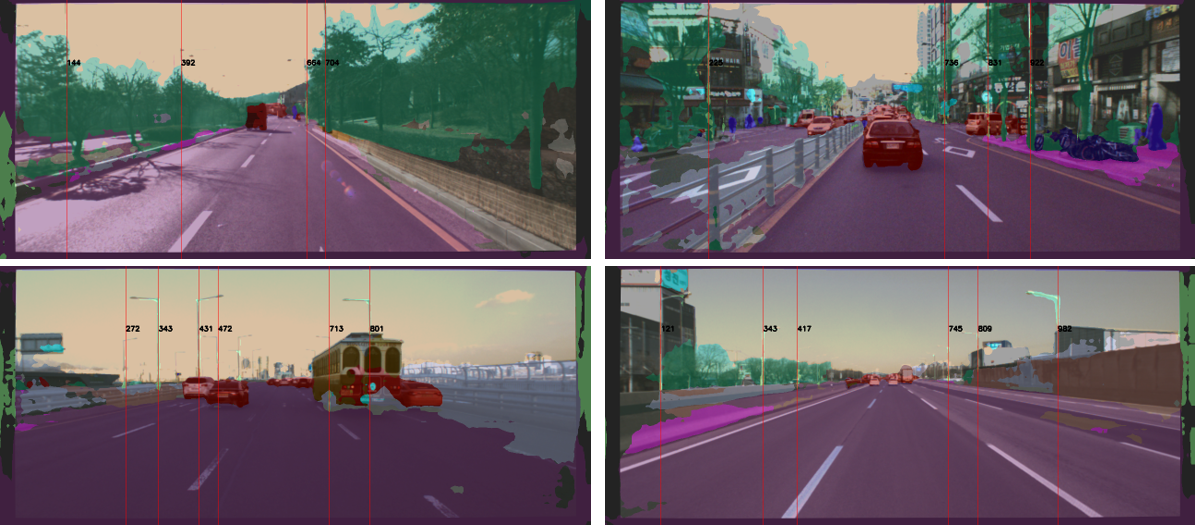}
                \end{center}
                \caption{Pole-extraction results of suburban, urban, bridge and highway, which are marked with red lines.}
                \label{fig:pole-extraction}
        \end{figure}
        
        We choose the KAIST Urban dataset \cite{jeong2019complex} to test our system in real data. KAIST provides detailed 3D lidar point clouds, stereo camera images, and wheel odometry data covering urban, suburban, highway, and bridge scenes, which is an ideal dataset to test how our pole-based system will work in common scenes for autonomous driving. Since the trajectories are too long and repeatable, we select Urban23, Urban26, and Urban34 from these sequences and divide continuous sections with clear poles into seven parts according to their scene types.
        
        We choose the VirtualKitti dataset \cite{Gaidon:Virtual:CVPR2016} to evaluate the performance of our system in synthetic datasets further. VirtualKitti provides the ground truth of semantic segmentation images and camera odometry. The ground truth of segmentation images can verify the accuracy limit of our proposed algorithm. Our experiments select continuous sections from trajectory 0001 that contain clear poles, whose timestamp range is 249-356. 
        
        In our experiments, we take four benchmarks altogether into consideration. RMSE Trans(m) and RMSE Rot(deg) are Root Mean Squared Error of camera translation and rotation. We also consider two benchmarks from \cite{stenborg2018long} and \cite{sattler2018benchmarking}. The former one, which we note as 0.5/1/2, counts the percentage of camera positions within 0.5/1/2 meters from the ground-truth. This benchmark can clearly illustrate whether the system is correct most of the time. However, it only considers the camera translation. Therefore, we use the latter benchmark, noted as 0.25/0.5/5, to count the percentage of camera positions within 0.25/0.5/5 meters and 2/5/10 degrees from the ground-truth.
        
        \subsection{Map Creation}
        
        This paper focuses on the localization models, but we will give a brief introduction of how we build the map.
        In both datasets, we manually annotate poles from point clouds to generate the compact map. In the KAIST dataset, the point clouds are given directly. As no point clouds are offered in the Virtual Kitti dataset, we use a structure-from-motion pipeline to build point clouds using ground-truth depth images.
        For the baselines' maps, we build them using the structure-from-motion pipeline as the same as \cite{stenborg2018long} for both datasets. 
        
        The map storage is shown in Table \ref{table:mapsize}. The SIFT descriptor needs large storage space, and the semantic point cloud requires a large number of points. Our compact map exceeds other maps in map storage significantly, which shows the apparent advantages of the object-level compact map.

        \begin{table}[!t]\footnotesize
                \begin{center}
                \caption{Comparison of Map Storage}
                
                {\begin{tabular}[l]{@{}lcccc}
                                
                                \toprule
                                
                                \multirow{2}{*}{Trajectory} & \multirow{2}{*}{Length} & \multicolumn{3}{c}{Map Storage} \\
                                \cmidrule(r){3-5}
                                & & UKF+SIFT & PC Semantic & Ours \\
                                
                                \midrule
                                KAIST26 & 3.7 km & 5.9 GB & 24.0 MB
                                & $\bm{10.0}$ \textbf{KB} \\
                                
                                Highway & 1.8 km & 2.1 GB & 8.2 MB
                                & $\bm{3.8}$ \textbf{KB} \\
                                
                                Bridge & 1.4 km & 3.3 GB & 5.0 MB
                                & $\bm{5.5}$ \textbf{KB} \\	
                                
                                VirtualKitti & 0.1 km & 36.8 MB & 0.14 MB & $\bm{0.7}$ \textbf{KB} \\

                                \bottomrule
                                
                \end{tabular}}

                \label{table:mapsize}
                \end{center}	
        \end{table}
        
        \subsection{KAIST Urban}
        
        %
        %
        %
                
        
        In the KAIST Urban dataset, we consider four typical scenes in autonomous driving tasks: urban, suburban, highway, and bridge. We select continuous sections in the dataset for each scene and compare our system with the baselines we construct. We also test how changes in the scene will influence the performance of our system. We use images generated by the left camera. And then we use the Bisenet \cite{yu2018bisenet} trained on Cityscapes \cite{cordts2016cityscapes} to segment the images.
        
\begin{figure}[t]
                \begin{center}
                        
                        \subfigure[Suburban]{
                                \begin{minipage}[t]{0.5\linewidth}
                                        \centering
                                        \includegraphics[width=1.0\linewidth]{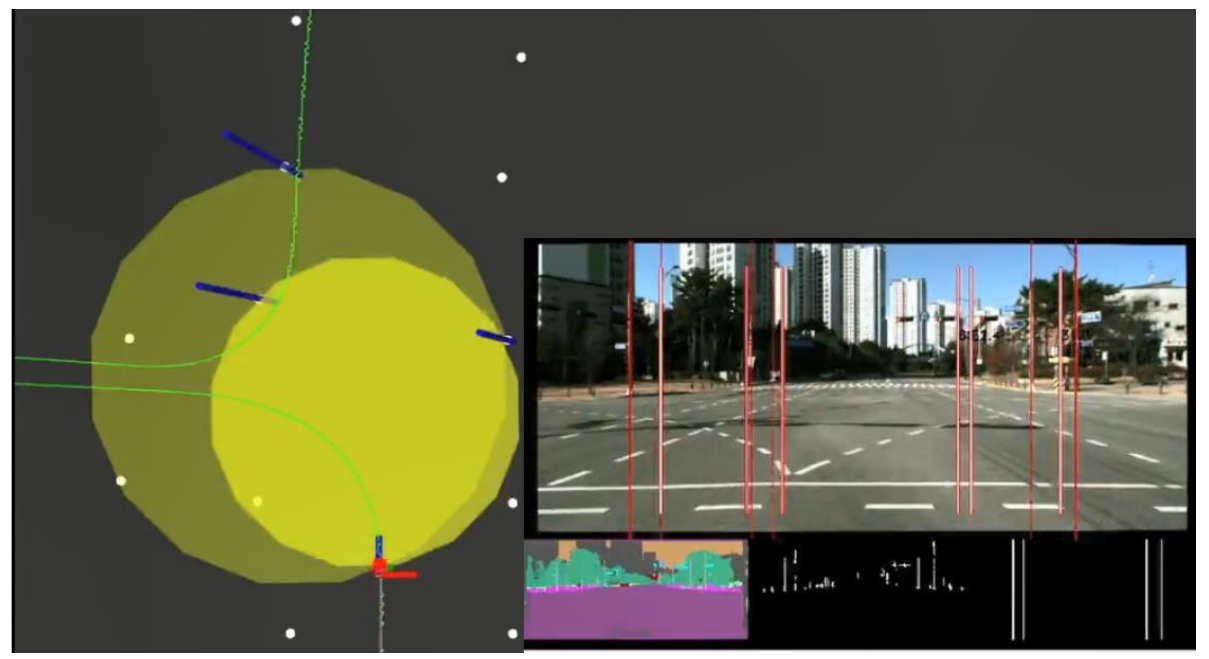}  
                                \end{minipage}%
                        }%
                        \subfigure[Urban]{
                                \begin{minipage}[t]{0.5\linewidth}
                                        \centering
                                        \includegraphics[width=1.0\linewidth]{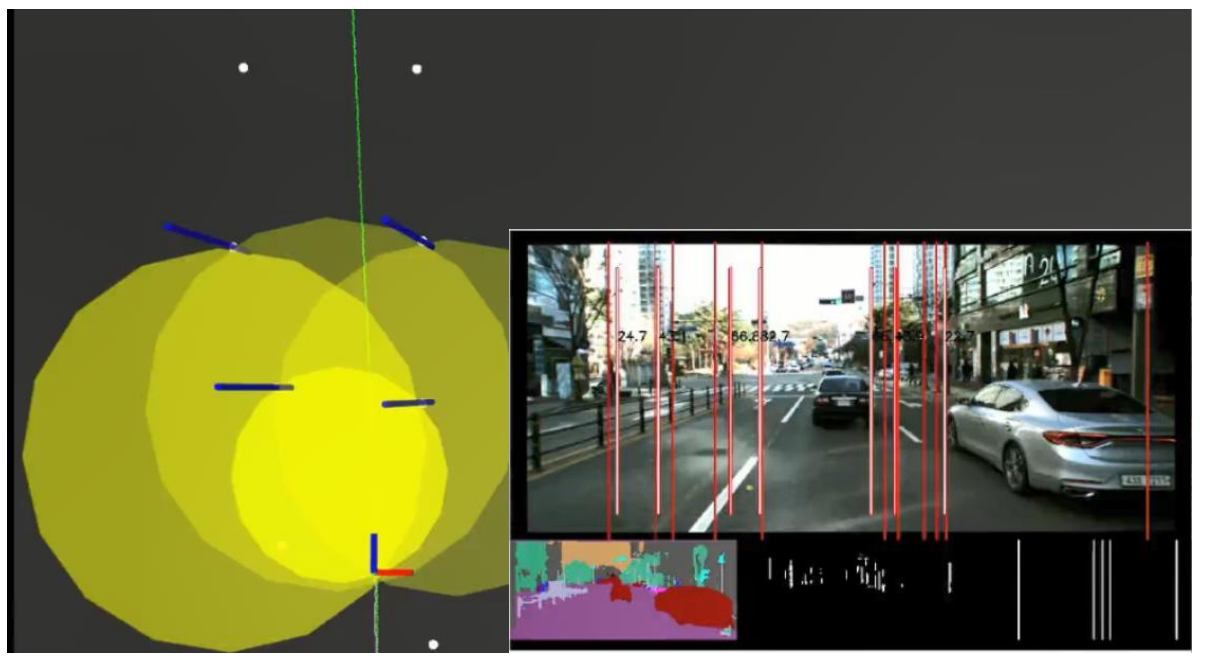} 
                                \end{minipage}%
                        }%
                        \quad
                        \subfigure[Highway]{
                                \begin{minipage}[t]{0.5\linewidth}
                                        \centering
                                        \includegraphics[width=1.0\linewidth]{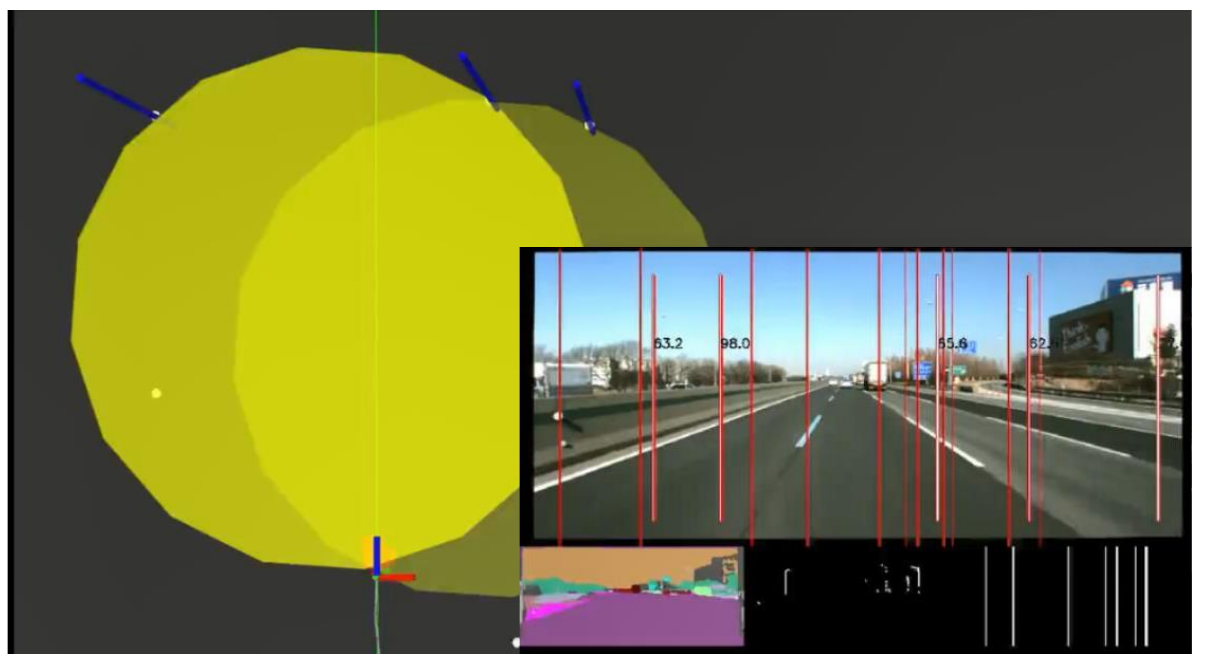} 
                                \end{minipage}%
                        }%
                        \subfigure[Bridge]{
                                \begin{minipage}[t]{0.5\linewidth}
                                        \centering
                                        \includegraphics[width=1.0\linewidth]{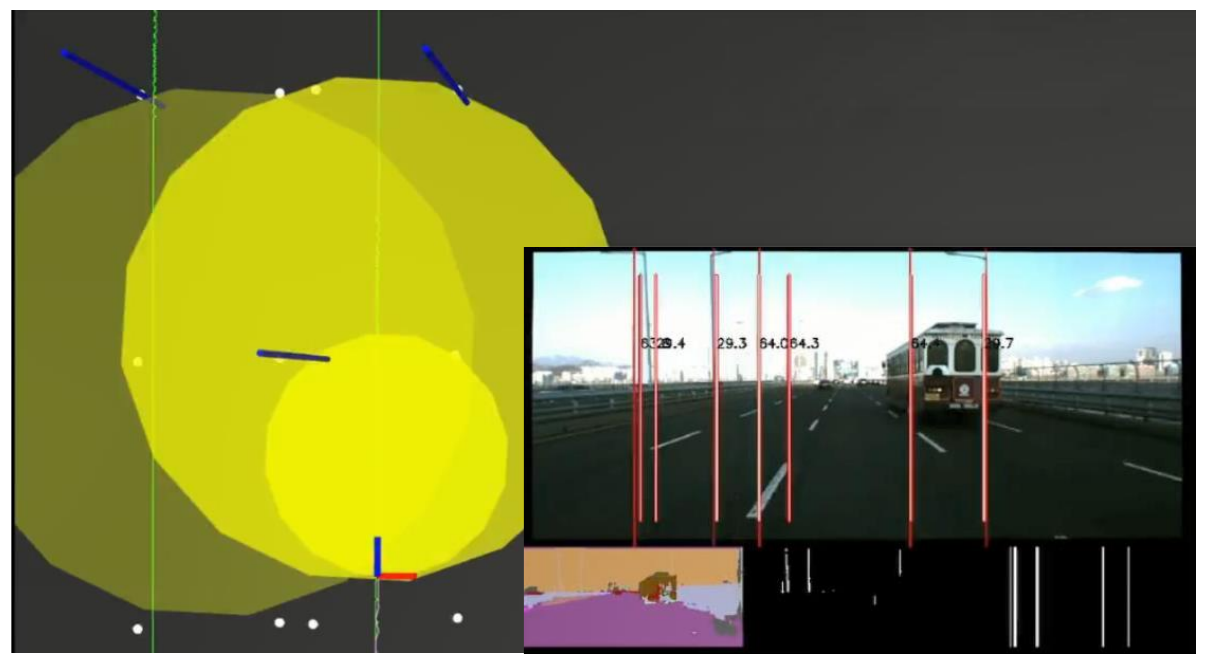}
                                \end{minipage}%
                        }%

                \end{center}
                \caption{Pose alignment module. The module solves the camera poses decoupling translation and rotation. The camera is located on the intersection of circles generated from pole-like objects observations. We recommend viewing the figure in color. We have attached a video to the supplementary material. }
                \label{fig:poseAlign}
                
        \end{figure}

\begin{figure}[h]
                \begin{center}
        
                        \includegraphics[width=0.9\linewidth]{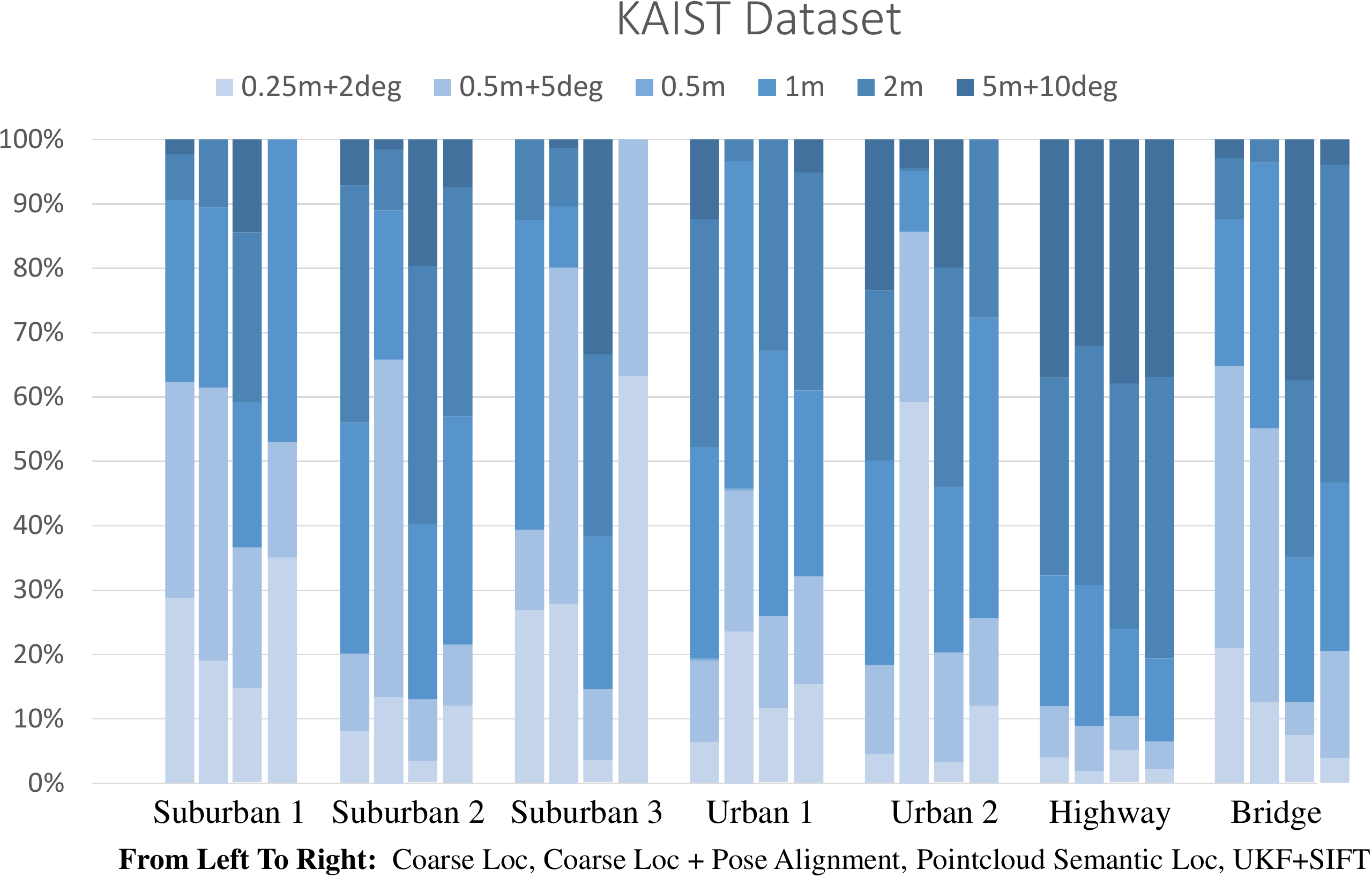}
                        ~\\
                        ~\\
                        \includegraphics[width=0.9\linewidth]{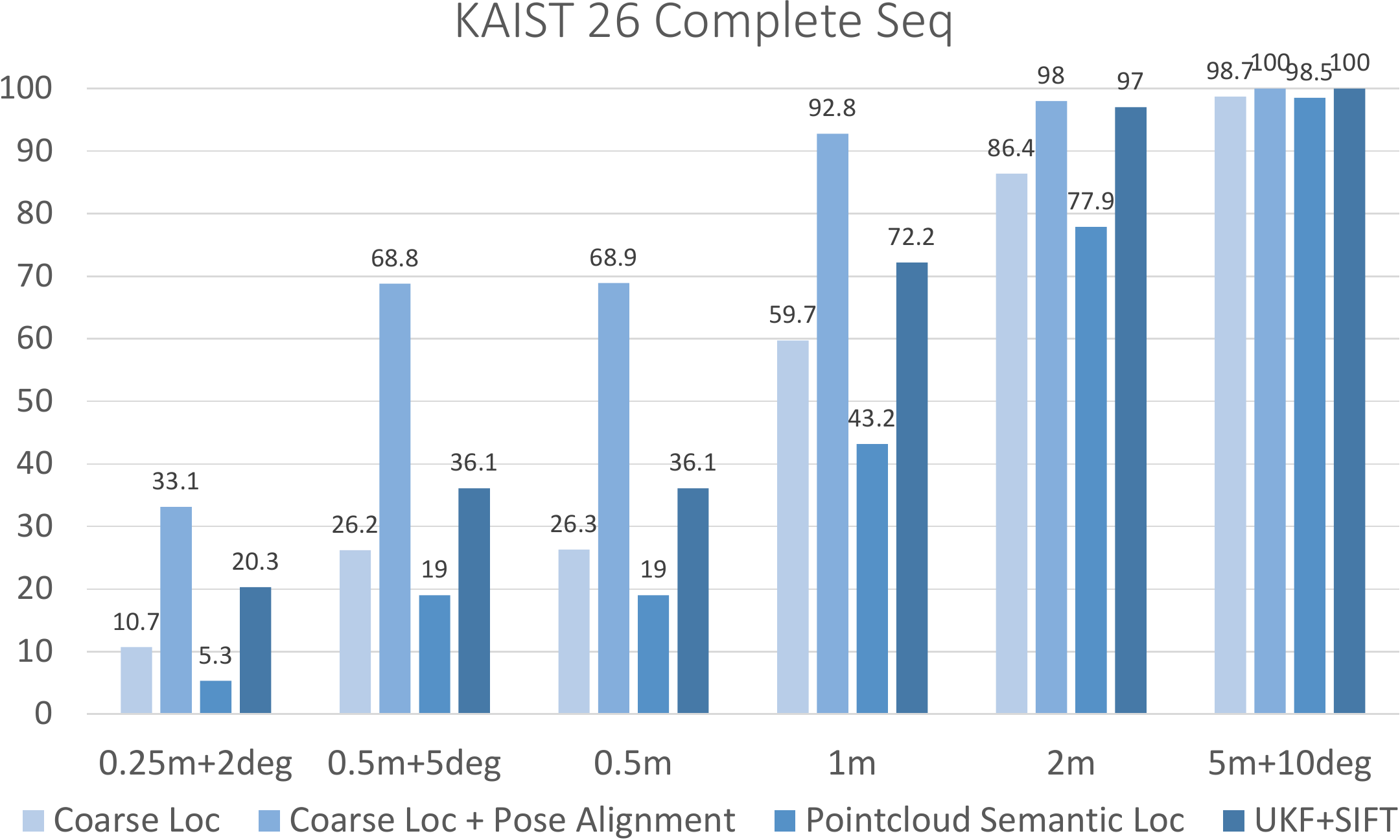}
                \end{center}
                \caption{Comparison of the KAIST dataset. The upper graph shows the results of 4 algorithms in 7 sequences. The bottom graph looks deeper into Kaist 26 sequence. Both graphs show that ours with pose alignment achieves a higher percentage of accurate localization.}
                \label{fig:kaist}
                
        \end{figure}

        The results in Table \ref{table-kaist-rmse} show that our system achieves comparable translation and rotation accuracy with the baselines, even though it uses a much more light-weighted map. When the Pose Alignment module is added, the translation accuracy is improved significantly and even exceeds the baselines on some trajectories. Some results of the Pose Alignment module are shown in Fig. \ref{fig:poseAlign}. We need to mention that on the KAIST26 trajectory, the map size of the UKF-SIFT system is 5.9 GB, while the map size of our semantic compact map is just 10.0 KB. Fig. \ref{fig:kaist} shows that the proposed system localizes the vehicle with an error below 1 m most of the time. 
        
        \begin{table}[!t]\footnotesize
                \caption{RMSE of KAIST dataset}
                \begin{center}
                {\begin{tabular}[l]{@{}lcccc}
                                
                                \toprule
                                & Coarse Loc & CL+PA
                                
                                & PC Semantic\cite{stenborg2018long} & UKF-SIFT  \\
                                \midrule
                                \multicolumn{5}{c}{Suburban 1}\\
                                Trans (m)& 0.686
                                & 0.604
                                
                                & 1.798
                                & $\bm{0.509}$
                                \\
                                
                                Rot (deg) & 0.882	
                                &  0.882
                                & 0.464
                                & $\bm{0.230}$
                                \\
                                
                                \midrule
                                
                                \multicolumn{5}{c}{Suburban 2}\\
                                Trans (m)& 1.344
                                & $\bm{0.674}$
                                & 2.034
                                & 1.088
                                \\
                                
                                Rot (deg) & 1.155
                                &  0.868
                                & 1.205
                                & $\bm{0.666}$
                                \\
                                
                                \midrule
                                
                                \multicolumn{5}{c}{Suburban 3}\\
                                Trans (m)&0.750
                                & 0.631
                                & 1.774
                                & $\bm{0.246}$
                                \\
                                
                                Rot (deg) & 0.841
                                &  0.760
                                & 1.350
                                & $\bm{0.407}$
                                \\
                                
                                \midrule
                                \multicolumn{5}{c}{Urban 1}\\
                                Trans (m)&1.589
                                & $\bm{0.580}$
                                
                                & 0.893
                                
                                & 1.051
                                
                                \\
                                
                                Rot (deg) & 1.195
                                &  1.080
                                
                                & 0.914
                                
                                & $\bm{0.562}$
                                
                                \\
                                \midrule
                                \multicolumn{5}{c}{Urban 2}\\
                                Trans (m)&1.955
                                
                                & $\bm{0.663}$
                                
                                & 1.669
                                
                                & 1.045
                                
                                \\
                                
                                Rot (deg) & 2.358
                                
                                &  0.838
                                
                                & 1.203
                                
                                & $\bm{0.622}$
                                
                                \\
                                
                                \midrule
                                \multicolumn{5}{c}{Highway}\\
                                Trans (m)&1.948
                                & $\bm{1.806}$
                                & 2.494
                                & 1.858
                                
                                \\
                                
                                Rot (deg) & 1.139
                                &  0.935
                                & 0.907
                                & $\bm{0.621}$
                                
                                \\
                                \midrule
                                \multicolumn{5}{c}{Bridge}\\
                                Trans (m)&0.737
                                &$\bm{0.593}$
                                &2.274
                                
                                & 1.139
                                
                                \\
                                
                                Rot (deg) & 1.031
                                &  0.845
                                & 1.577
                                
                                & $\bm{0.459}$
                                
                                \\
                                \midrule
                                \multicolumn{5}{c}{KAIST 26 Complete Seq*}\\
                                Trans (m)&1.539
                                
                                &$\bm{0.639}$
                                
                                &1.794

                                & 0.957
                                
                                \\
                                
                                Rot (deg) & 1.634
                                
                                &  0.902
                                
                                & 1.054

                                & $\bm{0.567}$
                                
                                \\
                                \bottomrule
                                
                \end{tabular}}
                \begin{tablenotes}
                        \footnotesize
                        \centering
                        \item[*]$^*$ KAIST 26 contains all the suburban and urban trajectories.
                \end{tablenotes}
                \label{table-kaist-rmse}
        \end{center}
        \end{table}

        \begin{figure}[!t]
                \begin{center}
                        \includegraphics[width=0.9\linewidth]{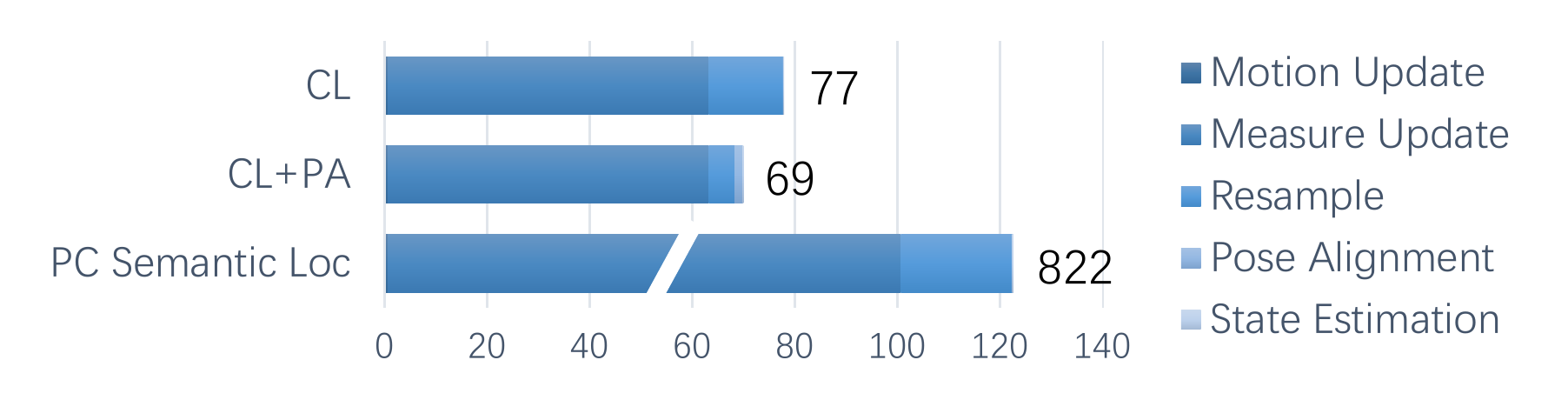}
                        
                \end{center}
                \caption{Comparison of run-times(ms). Ours save much time in Measure Update because of the compact map representation and landmarks number. The accurate resample costs less time than the resample in coarse localization, making CL+PA have the smallest average run-times.}
                \label{fig:time}
                
        \end{figure}

        In the bridge and highway scenes, the feature-based SIFT-UKF system lacks texture information. And it's hard for PC Semantic system to localize since there is little semantic difference in the longitudinal direction because of geometric configuration \cite{stenborg2018long}. However, poles are clear for our system as in Fig. \ref{fig:pole-extraction}.
        The results in the urban scenes demonstrate the robustness of our system in dynamic environments. Feature-based localization is susceptible to dynamic object interference. However, even under the occlusion of vehicles and people, the pole's height makes it still very clear in the image. 
        The proposed system performs not so well in suburban1 and suburban3. We find that the poles are mixed with vegetation, reducing the accuracy of the segmentation algorithm.
        
        The semantic baseline system performs not so well. One possible reason may come from the orientation of the camera. In the KAIST dataset, the camera points to the front of the vehicle, while Stenborg et al. \cite{stenborg2018long} point the camera at the sides of the vehicle, where more semantic diversity can be found. In our experiments, the localization accuracy quickly recovers when the vehicle turns.
        
        We also compare the time efficiency of our system with the semantic point cloud system. All the particle filter systems use 1000 particles. The average time consumption of each frame on the KAIST26 trajectory is presented in Fig. \ref{fig:time}. Our system is faster than the semantic point cloud system thanks to much fewer map points in the Measure Update process. We project all the poles in the map for the Measure Update. It is possible to get more time efficiency by maintaining a local map, which will be a valuable future work.
        
        \subsection{Virtual Kitti}
        
        Since no odometry data is given in the Virtual Kitti dataset, we add some noise to the ground-truth provided by the dataset and use it as the odometry for all three systems. As is mentioned by \cite{stenborg2018long}, such odometry is reasonable and can be used to simulate the motion model of autonomous vehicles. 
        
        \begin{table}[!t]\footnotesize
                \begin{center}
                \caption{RMSE of VIRTUAL KITTI}
                
                {\begin{tabular}[l]{@{}lcccc}
                                
                                \toprule
                                
                                & Coarse Loc & CL+PA
                                
                                & PC Semantic\cite{stenborg2018long} & UKF-SIFT  \\
                                
                                \midrule
                                Trans (m)& 0.346 & $\bm{0.289}$
                                & 0.326 & 0.302 \\
                                
                                Rot (deg) & 0.366
                                &  0.322
                                & 0.457
                                & $\bm{0.128}$
                                \\

                                \bottomrule
                                
                \end{tabular}}

                \label{table:vk}
                \end{center}
        \end{table}
        
        %
        %

        The results in Table \ref{table:vk} demonstrate that, even though we use much less information than the baselines, our system achieves comparable accuracy. Given the true segmentation result of the synthetic dataset, the translation accuracy with the pose alignment module can exceed the UKF-SIFT baseline and achieves an RMSE of 0.289m. The baselines use much bigger maps, as is shown in table \ref{table:mapsize}. 
        
        \subsection{Discussion}
        
        Failure cases in our experiments come from the following reasons, semantic segmentation accuracy and lack of observations. For example, barricades in the middle of roads are easy to be classified as poles. When the vehicle passes through sections with few poles, or the segmentation algorithm fails to segment enough poles, the error will accumulate since only the odometry is useful. These problems could be improved by improving the segmentation algorithm and expanding semantic objects from poles only into more types, like tree-trunks, traffic signs, etc. 
        
        For the rotation accuracy, we can tell that ours is better compared with the semantic baseline. However, there is still a certain gap compared with the UKF-SIFT baseline. The feature-based system has more observations to fuse. Since our geometric solver is based on the 2D assumptions, we believe it is a valuable future work to extend it to 3D space and verify the potential for improving rotation accuracy.
        
        
        \section{Conclusion}
        
        We propose a coarse-to-fine localization system based on pole-like objects extracted from semantically segmented images. The experiments demonstrated that even using a significantly small compact map, it is possible to achieve comparable accuracy with traditional point-cloud based localization. The pose alignment module decouples translation and rotation, achieving even better translation accuracy than the baselines. We prepare to explore more types of objects and their geometry representation to fill the conditions with no poles and explore 3D constraints to adapt 6-DOF state estimation in the future. 

\addtolength{\textheight}{-1cm}   



\section*{ACKNOWLEDGMENT}

This work was supported by the National Natural Science Foundation of China (No. 91748101). The authors want to thank Mr. Zidong Zhao and Ms. Zongyue Wang in Beihang University for the help in experiments.


\bibliographystyle{IEEEtran}
\bibliography{egbib}

\end{document}